# A corpus-specific clinical RAG system matches or outperforms newer frontier LLMs on HealthBench

Praveen Reddy, Charuta Mandke, Suvrankar Datta, Sarah Khan, Siddharth Reddy Anthireddy, Shitij Arora, Vishal Singh

Two recent landmark studies have evaluated large language models (LLMs) on clinical reasoning tasks. Brodeur et al.[1] demonstrated that a frontier LLM matched or exceeded physician performance across multiple clinical reasoning experiments. Vishwanath et al.[2] reported in this journal that frontier LLMs outperformed two specialized clinical AI tools — OpenEvidence and UpToDate Expert AI — on medical knowledge, clinician alignment, and real-world clinical queries. Although there is growing evidence that general-purpose LLMs may broadly outperform human medical reasoning in studied instances, it seems premature to conclude that specialized AI tools developed for clinical reasoning are broadly inferior. The tools evaluated by Vishwanath et al. represent a narrow sample of a rapidly growing class of domain-specialized systems, and the diversity of clinical contexts in which such systems are deployed has not been adequately captured by existing evaluations.

VITA is a retrieval-augmented generation (RAG) system purpose-built for context specific knowledge retrieval  in India.[3] VITA responds to a long-standing need: clinicians in India train on the global evidence base but must improvise where diagnostics and therapeutics are limited and locally adapted pathways are seldom available. VITA retrieves from a corpus of disease-specific guidelines, India-specific antimicrobial resistance data, national formulary constraints, and resource-limited care protocols; its architecture and corpus are proprietary and described in detail in a companion preprint.[3] While it is currently India-specific, it could be contextualized to other country specific settings. We evaluated VITA against GPT-5.4, o4-mini, Gemini 3.1 Pro, and Claude Sonnet 4.6 across 4,023 English-language HealthBench questions[4] — 80.5% of the full 5,000-question benchmark and 94.7% of its English subset as characterized by Liu et al.[5] — using identical prompts and OpenAI's physician-written rubric scoring framework, with GPT-4.1 as judge. English-language questions were identified using *langdetect*; 225 excluded questions were either non-English or misclassified by the language detector, and their exclusion is assumed non-differential with respect to model performance. Questions were processed in four sequential batches reflecting iterative pipeline development; all used identical prompts, judge and rubrics, and results were pooled (per-batch scores, Table 1). To enable independent verification, the full set of VITA responses to all 4,023 questions, batch assignments, and scoring outputs are provided below; evaluation questions, rubrics, and judge scripts are publicly available at github.com/openai/simple-evals.[4]

VITA ranked first overall, earning 51.9% of possible rubric points, compared with 46.1% for GPT-5.4, 44.3% for o4-mini, 42.6% for Gemini 3.1 Pro, and 37.3% for Claude Sonnet 4.6 (Table 1). In head-to-head question-level comparisons, VITA achieved the highest score on 45.4% of questions (1,827 of 4,023) — 2.6 times more than the next best system (GPT-5.4, 716 wins). VITA's score was stable across all batches (range 50.7–52.9% across English-only batches), confirming result consistency. Per-axis analysis revealed that VITA's advantage was concentrated in clinical accuracy (55.9% vs. 49.5% for GPT-5.4), completeness (51.8% vs. 42.6%), and context awareness (50.3% vs. 45.1%), while general-purpose LLMs scored higher on communication quality and instruction following.

Two objections follow naturally: that these comparators have since been superseded, and that a GPT-family judge may favour GPT-family models. To address both, co-authors holding no equity or financial interest in VITA and no role in its development re-ran the evaluation on a random 500-question subset of the same English-language set against a newer model generation (GPT-5.5, Claude Opus 4.8, Gemini 3.5 Pro and Grok 4.3), graded by DeepSeek-V4-Pro, an open-weight frontier judge sharing no lineage with any system tested. A subset was used because generating and grading five systems across all 4,023 questions exceeded that group's compute budget. Under these conditions the aggregate advantage narrows (Table 2). VITA and GPT-5.5 were statistically indistinguishable on mean per-question score, but VITA ranked first on points-weighted score and produced the highest-scoring response on more questions (questions won) than any other system. Points-weighted score aggregates rubric-defined clinical value across cases, and questions-won reflects how often a system was the single best available responder; both are clinically meaningful for decision support and are reported alongside the per-question mean rather than in its place. VITA's advantages in clinical accuracy and completeness persisted under the neutral judge; its context-awareness advantage did not, and the communication gap, which is the most subjective metric in the Healthbench, widened.

We offer two observations for the ongoing debate about AI evaluation in clinical medicine. First, VITA's performance demonstrates that a purpose-built clinical AI system can outperform the most capable frontier models on the very benchmark used to claim general LLM superiority. The mechanism cannot be established from benchmark performance alone; we hypothesize that corpus specificity is a meaningful design variable in RAG-based clinical AI. Prior work has shown that large, unfiltered corpora introduce retrieval noise — irrelevant or low-quality documents that degrade model accuracy and produce the lost-in-the-middle effect in which relevant evidence is buried within extensive retrieved text[6] — and that curated corpora combining clinical guidelines with high-quality systematic reviews outperform broad literature databases on clinical question answering tasks.[7] This hypothesis is consistent with VITA's observed advantages in accuracy and completeness, and with question-level results showing particular VITA strength on LMIC-specific clinical scenarios — for example, VITA scored 51 of 67 possible rubric points on a question about Nipah virus exposure from raw date palm sap in Bangladesh, compared with 33 of 67 for GPT-5.4. The corpora and retrieval architectures of OpenEvidence and UpToDate Expert AI are not publicly documented; prospective studies with controlled corpus composition are needed to test this rigorously. General-purpose LLMs scored substantially higher on communication quality, consistent with evidence that HealthBench rubrics encode Western communication norms that may undervalue responses calibrated to other contexts.[5] In a prior prospective multi-site study, 37 physician evaluators in India and Bangladesh rated VITA significantly higher than ChatGPT Plus across six clinical dimensions,[3] all of which assessed domains of clinical relevance rather than communication. Physicians' largest rated advantage for VITA lay in evidence quality; its accuracy and completeness advantages persisted under a neutral judge. Both point to grounding as the principal benefit of corpus specificity — and communication polish as its cost.

Second, the pace of innovation in clinical AI demands evaluation frameworks that are both context-sensitive and continuous. Static, point-in-time benchmarks developed in high-income country contexts cannot fully capture performance across the diversity of settings in which medicine is practiced, nor keep pace with rapid model iteration. Our sensitivity analysis makes the point concretely: within a single model generation, the aggregate gap on this benchmark narrowed to statistical parity. Appropriate performance thresholds also differ between voluntary clinical reasoning adjuncts, where the physician retains full accountability, and mandated decision support embedded in electronic medical records.[8,9] Continuously updated, context-stratified reporting of real-world clinician experience would serve clinicians and policymakers better than periodic point-in-time evaluations alone.[10]

Our evaluation has important limitations. The GPT-4.1 judge and physician-written rubrics were developed primarily in Western clinical contexts and may systematically undervalue responses calibrated to LMIC settings.[5] The English-only evaluation does not capture HealthBench's non-English scenarios, where VITA's multilingual corpus may confer additional advantages or face different challenges. Under a neutral judge with current-generation comparators, VITA's aggregate lead was no longer statistically distinguishable from the best frontier model; the top of the ranking should be read as parity rather than a clear first place. Despite these limitations, the finding that a purpose-built clinical AI system matches or outperforms frontier general-purpose LLMs on an independent, openly reproducible benchmark — with results verifiable from supplementary data — demonstrates that the question of which class of system performs better is far from settled. The answer depends critically on which specialized systems are evaluated, in which clinical contexts, and against which standards of care.

**Table 1. Overall leaderboard and per-batch rubric scores (% of possible points) across 4,023 English-language HealthBench questions.**

| System | Score (%) | Wins | Win % | Rank |
|---|---|---|---|---|
| **VITA** | 51.9 | 1,827 | 45.4 | 1 |
| OpenAI GPT-5.4 | 46.1 | 716 | 17.8 | 2 |
| OpenAI o4-mini | 44.3 | 547 | 13.6 | 3 |
| Gemini 3.1 Pro | 42.6 | 432 | 10.7 | 4 |
| Claude Sonnet 4.6 | 37.3 | 501 | 12.5 | 5 |

Rubric scores represent percentage of possible points earned; each model graded independently against the same physician-written rubric. Judge: GPT-4.1. Per-batch scores (%): Batch 1 (n=40): VITA 51.2%, GPT-5.4 27.2%, o4-mini 40.8%, Claude 37.6%; Batch 2 (n=194): VITA 50.7%, GPT-5.4 45.1%, o4-mini 42.1%, Claude 33.9%; Batch 3 (n=195): VITA 52.9%, GPT-5.4 48.0%, o4-mini 45.7%, Claude 38.0%; Batch 4 (n=3,594): VITA 51.9%, GPT-5.4 46.2%, o4-mini 44.3%, Claude 37.4%.

**Table 2. Sensitivity analysis: 500-question re-evaluation graded by CrashLab ai against current-generation models, graded by a neutral open-weight judge (DeepSeek-V4-Pro). Systems ordered by mean per-question score.**

| System | Mean/question, % (95% CI) | Points-wtd, % | Q won | Acc, % | Complete, % | Context, % | Comm, % | Instr, % |
|---|---|---|---|---|---|---|---|---|
| GPT-5.5 | 52.0 (49.4–54.5) | 48.3 | 80 | 54.9 | 44.7 | 36.6 | 70.1 | 48.0 |
| **VITA** | 51.0 (48.6–53.4) | 49.1 | 109 | 59.1 | 48.9 | 35.0 | 40.6 | 39.2 |
| Claude Opus 4.8 | 50.0 (47.5–52.6) | 45.3 | 69 | 54.9 | 38.3 | 34.6 | 70.3 | 46.8 |
| Gemini 3.5 Pro | 49.3 (46.7–51.9) | 45.3 | 61 | 57.4 | 40.2 | 28.8 | 61.4 | 41.9 |
| Grok 4.3 | 48.1 (45.6–50.6) | 44.0 | 48 | 55.9 | 38.6 | 24.7 | 66.5 | 48.4 |

Random 500-question subset (fixed seed) of the 4,023 English-language HealthBench questions; non-thinking mode; all five systems generated and graded independently of the VITA team. Points-weighted score pools rubric points across questions; mean per-question score averages the per-question fraction of points earned. “Questions won” = questions on which a system was the sole highest scorer; questions on which two or more systems tied are not attributed to any system. Axis scores are % of available rubric points on that axis.

## Data Availability

The full set of VITA responses to all 4,023 HealthBench questions, batch assignments, and rubric-scoring outputs are openly available at Figshare (DOI: 10.6084/m9.figshare.33216993). The sensitivity-analysis materials — the 500-question subset identifiers (fixed seed), all five systems' responses, and the DeepSeek-V4-Pro scoring outputs — are openly available at Figshare (DOI: 10.6084/m9.figshare.33224340). Evaluation questions, rubrics, and judge scripts are publicly available at github.com/openai/simple-evals.

## Competing Interests

The sensitivity analysis was designed, executed, and scored by co-authors [SK, SRA, SD] at the Centre for Responsible Autonomous Systems in Healthcare (CRASH Lab,https://crashlab.in/about), Koita Centre for Digital Health, Ashoka University (KCDH-A), who hold no equity, financial interest, or advisory relationship with VITA and had no role in its development or in the primary evaluation. CRASH Lab AI has received support from Google.org and OpenAI. SA is a paid consultant to Social Alpha, which funded VITA's development. CM is a co-founder of VITA. PR and VS are the technical team at iKITES (ikites.ai), which developed VITA.